\documentclass[runningheads]{llncs}
\usepackage[T1]{fontenc}

\usepackage[hyphens]{url}  
\usepackage{graphicx} 
\usepackage{xcolor}
\usepackage{booktabs}
\usepackage{siunitx} 
\usepackage{pgfplotstable} 
\pgfplotsset{compat=1.18}
\usepackage{subcaption}
\usepackage{csvsimple} 
\usepackage{multirow}
\RequirePackage[np]{numprint}

\usepackage[numbers,sort]{natbib} 
\newcommand{\citepos}[1]{\citeauthor{#1}'s [\citeyear{#1}]}
\DeclareRobustCommand{\nUmErAL}[1]{}

\usepackage[normalem]{ulem}
\newcommand{\rep}[2]{{\textcolor{green!50!black}{\sout{#1}}}{\textcolor{blue!50!black}{#2}}}

\usepackage[most]{tcolorbox}  
\usepackage{tikz,lipsum,lmodern}
\usepackage{setspace} 
\tcbset{
sharp corners,
colback = white,
coltitle=black,
before skip = 0.2cm,    
after skip = 0.5cm,      
boxrule=0.1pt
}    

\newtcolorbox{chatbox}[1]{
colback=gray!5!white,
colframe=gray!50!black,
colbacktitle=gray!25,
coltitle=black,
title={\vspace{0.1cm}\sffamily #1\vspace{0.1cm}},
rounded corners,
arc=0.8mm,
breakable
}

\DeclareRobustCommand{\newline}[1]{\\}
\DeclareRobustCommand{\expert}[1]{\includegraphics[scale=0.045]{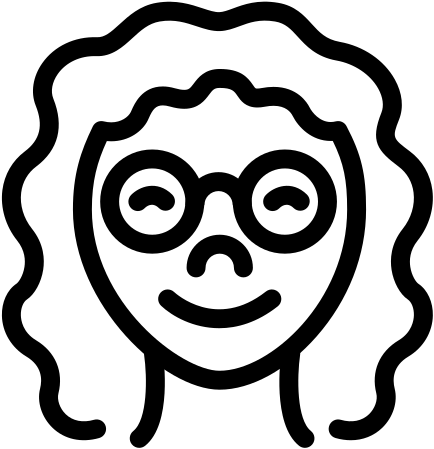} \emph{#1}\vspace{0.1cm}}
\DeclareRobustCommand{\prompt}[1]{\emph{#1}}
\DeclareRobustCommand{\gpt}[1]{\includegraphics[scale=0.058]{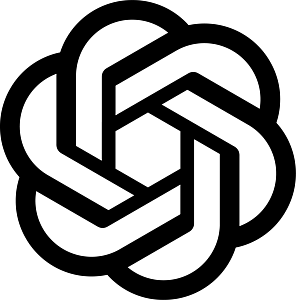} \emph{#1}\vspace{0.0cm}}

\begin{document}

\title{Extracting Norms from Contracts Via ChatGPT: Opportunities and Challenges}

\titlerunning{Extracting Norms from Contracts Via ChatGPT}

\author{%
Amanul Haque \and 
Munindar P. Singh
}%
\institute{
North Carolina State University\\
\email{ahaque2@ncsu.edu, mpsingh@ncsu.edu}  
}


\newcommand{\mps}[1]{\textcolor{blue!90!black}{\sffamily MPS:~~#1}}
\newcommand{\ah}[1]{\textcolor{red!90!black}{AH:~~#1}}

\maketitle

\begin{abstract}

We investigate the effectiveness of ChatGPT in extracting norms from contracts. 
Norms provide a natural way to engineer multiagent systems by capturing how to govern the interactions between two or more autonomous parties. 
We extract norms of commitment, prohibition, authorization, and power, along with associated norm elements (the parties involved, antecedents, and consequents) from contracts. 
Our investigation reveals ChatGPT's effectiveness and limitations in norm extraction from contracts.
ChatGPT demonstrates promising performance in norm extraction without requiring training or fine-tuning, thus obviating the need for annotated data, which is not generally available in this domain. 
However, we found some limitations of ChatGPT in extracting these norms that lead to incorrect norm extractions.
The limitations include oversight of crucial details, hallucination, incorrect parsing of conjunctions, and empty norm elements. 
Enhanced norm extraction from contracts can foster the development of more transparent and trustworthy formal agent interaction specifications, thereby contributing to the improvement of multiagent systems. 



\end{abstract}

\section{Introduction}

\emph{Normative relationships}---\emph{norms}, for short---typically refer to a social construct that governs the behavior of individuals in a given social context.
The behaviors could range from etiquette and social expectations to moral and ethical standards, depending on the context.
These norms help capture the essential aspects of interactions between individuals independent of low-level implementation details. 
Recent approaches have shown how we might engineer a multiagent system starting from specifying norms between member agents \cite{IC-21:Hercule}. 


Contracts are a rich source of norms.
Although contracts are expressed in natural language, it is well recognized that they can be understood as capturing the norms between the contracting parties \cite{gao-2014AAMAS:extracting_norms_from_contracts}.
Norms in contracts define the terms of the agreement between the contracting parties and their expected behaviors. 
This paper shows how we might begin from existing annotated contracts to identify relevant norms. 
Even if we don't fully formalize the norms in this way, progress made in identifying the norms can help in producing valid norm specifications for further implementation.  


We use ChatGPT, a pretrained Large Language Model (LLM), to extract norms from contracts.
LLMs are sophisticated AI systems capable of understanding language and have pushed the state-of-the-art on many standard language benchmark tasks \cite{radford-OpenAI2018:improving_language_understanding, devlin-2019NACL:bert, liu-2019arXiv:roberta, yang-2019NEURIPS:xlnet}.
We use a model of norms based on Custard \cite{AAMAS-16:Custard} and its variants, e.g., \cite{Computer-20:Violable}, and extract norms such as commitments, prohibitions, authorizations, and powers and conduct a qualitative analysis of the extracted norms.

Whereas ChatGPT demonstrates notable effectiveness in norm extraction, our analysis uncovers recurring challenges. 
Some recurring problems include overlooking crucial details, hallucinations, incorrect parsing of conjunctions, and empty norm elements in the extracted norms.
Additionally, our investigation emphasizes the scarcity of high-quality annotated datasets crucial for furthering norm extraction from contracts and, hence, advancement toward automated contract comprehension. 
Existing datasets are marred by limitations, including a small number of manually annotated contracts, lack of diversity in types of contracts and clauses, and poor manual annotations. 

Norms offer a principled basis for engineering flexible, autonomous multiagent systems for business and other cross-organizational settings.
Automating norm extraction from unstructured text documents can help us design more robust and efficient decentralized autonomous multiagent systems.  
A clear path to engineering a multiagent system based on norms derived from contracts can improve the comprehensibility of the resulting specifications and clearer accountability and, thus, trust.
Our efforts are a step forward in that direction. 
The insights from this study can guide future research aimed at overcoming barriers to effective automated norm extraction from natural text. 




The rest of this paper is organized as follows. 
Section~\ref{sec:Rlated} discusses related work. 
Section~\ref{sec:Methodology} describes the methodology (including the definitions, dataset, and the ChatGPT Prompt used). 
Section~\ref{sec:Findings} discusses the findings of our qualitative analysis, followed by a conclusion in Section~\ref{sec:Conclusion}.

\section{Related Work}
\label{sec:Rlated}

Automating comprehension of legal documents has long been a topic of investigation. 
Many prior works have automated the extraction of useful information from unstructured legal documents. 
This includes extracting software requirements \cite{sainani-2020IEEE-IREC:extracting_requirements_from_contracts}, unfair or risky clauses \cite{lagioia-2017LKIS:detect_unfair_clauses_in_online_contract,lee-2019JCCE:extract_poisonous_clauses_from_construction_contracts}, legal norms \cite{hashmi-2015-IEEE-W:extracting_legal_norms_from_contracts}, conflicting norms \cite{aires-2021-SIP:norm_conflict_identification}, and business events and their temporal constraints \cite{gao-IEEE-TSC:mining_buness_events_from_contracts}. 
However, despite considerable progress, the goal of achieving comprehensive automation in contract understanding is far from complete.

Automating contract execution from unstructured text documents involves two fundamental steps. 
\emph{First}, identifying the contract clauses and then inferring the relationship between the identified clauses. 
This involves identifying relevant text spans in the document (i.e., context) and extracting relationships between different entities from the context. 
\emph{Second}, translating the identified relationships into formal specifications suitable for computational processing, such as Hercule \cite{IC-21:Hercule} and Symboleo \cite{sharifi-2020IEEE-RE:symboleo}. 
In this work, we focus on the first task, more precisely, extracting norms (normative relationships) from unstructured legal contracts. 

A major roadblock in automating contract understanding is the lack of manually annotated gold-standard datasets. 
The manual annotation of contracts has long posed a challenge essentially due to greater complexity and intricacy of language in contract documents that require experts for annotations.
Additionally, contract documents are extensive in length (a contract may be a few pages to hundreds and sometimes even thousands of pages long). 
Although some previous works have created manually annotated contract datasets, they are often small (i.e., contain few manually annotated contracts), lack coverage (i.e., identify a limited number of different types of clauses and contracts), and exhibit poor quality human annotations. 

\citet{leivaditi-2020:benchmark_for_lease_contracts} annotate 179 lease agreements for named entities and red flags. 
Although their dataset is useful for automating contract comprehension, it focuses exclusively on one type of contract (leases) and a small number of types of clauses. 
\citet{koreeda-manning-2021EMNLP-findings:contractnli-dataset} create ContractNLI, a dataset for Natural Language Inference (NLI) on contracts. 
Given a contract and a set of hypotheses, they annotate each hypothesis as entailed by, contradicting, or not mentioned by (neutral to) the contract. 
They also identify evidence for the annotation as text spans in the contract. 
Although ContractNLI offers more manually annotated contracts ($\approx$\np{600}) than most datasets containing annotated contracts, its annotations are less granular. 
For instance, ContractNLI doesn't identify different contract clauses, which is essential for automated contract understanding.

For our analysis, we chose CUAD (Contract Understanding Atticus Dataset)
\cite{hendrycks-2021arXiv:CUAD}, a Contract Clause Extraction (CCE) dataset containing more than 500 contracts manually annotated by experts (lawyers and law students). 
CUAD proposes the task of contract comprehension as a question-answering task and contains text spans of the contract clauses manually identified as belonging to one of 41 categories. 
CUAD is more diverse than previous datasets on contracts (i.e., contains contracts from different domains), has a higher number of annotated contracts, and is annotated by experts. 
Further, the ontology is developed for the given dataset (as opposed to adapting an ontology from another work that may or may not be suitable for the current dataset).


Prior research has proposed several methodologies for extracting essential clauses from unstructured legal documents, including Case-Based Reasoning (CBR) \cite{agostini-1999ICAIL:JurisConsulto}, grammars, word lists, and heuristics \cite{nadzeya-2008CM-ER:extracting_rights_and_obligations_from_regulatory_compliance}, rule-based NLP \cite{sleimi-2021ESE:extracting_semantic_legal_metadata}, hand-crafted rules and machine learning \cite{curtotti-2010ALTA:corpus_of_australian_contracts}, neural models such as hierarchal BiLSTM with attention and CNN (Convolution Neural Networks) \cite{chalkidis-2018-ACL-Short:obligation_and_prohibition_extraction,chalkidis-2021CoRR:neural_contract_extraction}, and fine-tuned transformer-based models, such as Legal-BERT \cite{mamakas-2022NLLP:legal_bert}. 
Although some of the prior approaches are effective, they usually require a lot of manual effort, either in hand-crafting rules or annotating large amounts of data for training.

Recent advances in NLI and NLU (Natural Language Understanding) present promising avenues for alleviating the burden of manual effort in automated contract understanding. 
\citet{vaswani-2017NEURIPS:transformers} proposed Transformers, a deep learning model architecture based on multihead self-attention, which pushed the state-of-the-art performance for multiple NLI and NLU benchmarks, including the General Language Understanding Evaluation (GLUE) \cite{wang-2018EMNLP:glue_benchmark}, Stanford Question Answering (SQuAD) \cite{rajpurkar-2016EMNLP:squad_v1,rajpurkar-2018ACL-S:squad_v2}, and Multigenre NLI (MNLI) \cite{williams-2018ACL:mnli_corpus}. 
We use ChatGPT (model GPT-3.5-turbo), an instance of the Generative Pretrained Transformer (GPT), to extract norms from contracts.
ChatGPT has pushed the state of the art on multiple benchmarking language tasks.
ChatGPT showcases capabilities that closely mirror human-level performance on language understanding and inference and occasionally exceed human proficiency on specific tasks \cite{bubeck-2023arXiv:sparks_of_agi_exp_with_gpt4}.
ChatGPT has been used in several domains including healthcare \cite{malik-2023MDPI:chatgpt_in_healthcare, ali-2023LDH:chatgpt_for_patient_letters, howard-2023:chatgpt_and_antimicrobial_advice, khan-2023PJMS:chatgpt_reshaping_med_edu_and_management}, business and finance \cite{alafnan-2023JAIT:ChatGPT_for_business_writing, george-2023PUIIJ:chatgpt_impact_on_business}, creative content generation \cite{cox-2023:chatgpt_implications_for_academic_libraries, taecharungroj-2023bdcc:what_can_chatgpt_do}, writing code \cite{Kashefi-2023JMLMC:chatgpt_for_prgramming, jalil-2023IEEE-ICSTVVW:chatgpt_and_software_testing_edu}, and education and training \cite{baidoo_anu-2023JAI:chatgpt_in_promoting_teaching_and_learning, kasneci-2023LID:llms_for_edu}.

Although transformer-based models show promising results on a variety of NLU and NLI tasks, evidence suggests that most LLMs perform poorly in contract understanding \cite{xu-2022EMNLP:conreader}.
Further, there is a lack of qualitative analysis of how well the LLMs' perform in contract understanding. 
Such analysis is essential to devise reliable automated contract understanding systems that could streamline the process of contract analysis and review, saving valuable time and resources for legal professionals and organizations. 
To facilitate such efforts, we qualitatively evaluate the effectiveness of ChatGPT in understanding contracts via norm extraction.




\section{Methodology}
\label{sec:Methodology}

We extract norms (normative relationships), for short---from contracts (unstructured text documents) to facilitate the engineering of agents and multiagent systems that respect stakeholder needs as captured in contracts. 

In abstract terms, a norm creates a legitimate expectation of one autonomous party over another autonomous party. 
The sense of \emph{norm} used here is the one established in the legal literature \cite{Von-Wright-63:Norm} and is formal and prescriptive. 
A norm can be viewed as an element of a contract. Or, conversely, a contract is (primarily) a set of norms. 
Norms in contracts serve multiple purposes and fall into a variety of types of clauses \cite{IC-Ethics-21:graybox, hendrycks-2021arXiv:CUAD}. 

Norms provide a natural way to engineer multiagent systems because they provide a natural basis (i.e., one that is close to stakeholders' conception) for governance \cite{TIST-13-Governance,TOIT-22:governance} and accountability \cite{IC-21:accountability} in settings involving two or more autonomous entities. 
The benefits of using norms include improvements in trustworthiness and resilience of sociotechnical systems \cite{TOSEM-20:Desen,JAAMAS-23:ReNo}. 

We adopt \citepos{gao-2014AAMAS:extracting_norms_from_contracts} approach to define normative relationships to extract from a contract. 
A norm is directed from a subject to an object and is constructed as a conditional relationship involving an antecedent (which brings the norm into force, i.e., the condition on which the action of the subject depends) and a consequent (which brings the norm to satisfaction, i.e., the outcomes of the action). 

A norm can be one of the following four types: 
\begin{description}
\item[A \emph{commitment}] means that its subject commits to its object to ensure the consequent if the antecedent holds. 
\item[A \emph{prohibition}] means that its subject is forbidden by its object from bringing about the consequent if the antecedent holds. 
\item[An \emph{authorization}] means that its object authorizes its subject to bring about the consequent if the antecedent holds. 
\item[A \emph{power}] means that its subject is empowered by its object to bring about the consequent if the antecedent holds. 
\end{description}

The norm types of authorization and power involve similar language; however, some differences exist \citep{gao-2014AAMAS:extracting_norms_from_contracts}, which we can use to differentiate them. 
An authorization is when the subject is allowed to bring about the consequent of the norm. 
A power is when the subject brings about the consequent as a way to change an existing normative relationship. 
Consider a manufacturing contract. 
The manufacturer may use some equipment provided by the purchaser under the stated conditions. 
No norms change as a result of such usage. 
Therefore, the use of equipment maps most naturally to an authorization. 
The purchaser may cancel an order with prior notice, i.e., it can terminate a commitment at will, thereby changing the normative relationship between itself and the manufacturer. 
Therefore, termination in this case (and usually) is mapped to power. 

\subsection{Dataset}

There is a lack of good-quality gold standard datasets for contract understanding. 
This is partly because of the complexity of the task itself; annotating contracts often requires expert judgment. 
Further, there is no limit to the variety of clauses that may be present in a contract depending on its purpose. 
For instance, a manufacturing contract may have clauses that are not found in a lease agreement, and vice versa. 
Similarly, for other types of contracts. 
Not surprisingly, prior works focus on a few clauses to annotate depending on the task. 

For our analysis, we chose CUAD (Contract Understanding Atticus Dataset) that contains 510 manually annotated contracts.
To ensure diversity in the dataset, CUAD includes 25 types of contracts, such as manufacturing, service, and leasing. 
Contracts in CUAD have varying lengths, ranging from a few pages to over one hundred pages.
Each contract in CUAD is annotated manually by legal experts by identifying essential clauses (i.e., text spans in the contracts) that facilitate contract comprehension.
CUAD defines 41 types of clauses (categories), including \emph{temporal clauses} such as Effective Date, Renewal Date, and Expiration Date, \emph{service-related clauses} such as Governing Laws, Revenue-Profit Sharing, and Price Restrictions, and \emph{termination-related clauses} such as Termination for Convenience and Post Termination Services.
CUAD contains \np{13000} of such clauses identified manually by experts.
In addition to identifying the relevant spans, CUAD contains a short answer (derived from the relevant text spans) to a question for each clause, such as `\emph{on what date will the contract's initial term expire?}' (Expiration Date), or \emph{`Is one party required to share revenue or profit with the counter party for any technology, goods, or services?'} (Revenue/Profit Sharing). 

Although CUAD annotations are claimed to be reliable and double-checked by experts, we found instances where the annotations were lacking. 
We found some annotations leave out contexts (text spans from the contract) that may help answer the clause related questions more accurately. 
We manually verified some contracts and found termination-related clauses, such as Post-Termination Services and Termination For Convenience, often overlooked relevant text spans casting doubt on the quality of these annotations.
This shows the inherent difficulties of the task that is challenging even for experts. 
This may be due to long and complex sentence structures common in contracts or information relevant to one clause often being spread across the document, with each contract document's length ranging from a few pages to hundreds of pages.

Some clause types in CUAD are easy to identify.
For instance, temporal clauses such as \emph{agreement date} and \emph{expiration dates} have a distinctive syntax. 
Likewise, the \emph{parties} involved in a contract are easy to identify as they are explicitly mentioned at the beginning of the contract and repeated multiple times across the document.
For simplicity, we focus our analysis on clause types related to the termination of a contract, such as \emph{Termination For Convenience}, \emph{Right of first refusal} (or \emph{offer} or \emph{negotiation}) (Rofr/Rofo/Rofn), and \emph{Post-Termination Services}, and clause types related to the change of control of a party to the contract, such as \emph{Change of Control}. 
These clause types are a good venue to look for norms (normative relationships). 
Further, these are some of the toughest clause types to identify, as evidenced by the fact that three of these four clauses types were in the top six worst-performing clause types benchmarked using a Transformer model (DeBERTa) \cite{hendrycks-2021arXiv:CUAD}. 
We pick clauses from the CUAD dataset of the above-mentioned types and analyze the norms identified by ChatGPT. 
We randomly sample 100 clauses to evaluate.
Further, we identify 50 more clauses that potentially contain more challenging logical forms to extract, such as conditional actions and multiple conjunctions (OR, AND). 
We use keywords such as `and', `or', `upon', `unless', and so on to identify such challenging clauses.
We use ChatGPT to extract norms from these selected clauses.  

\subsection{Extracting Norms via ChatGPT}

ChatGPT is a state-of-the-art LLM based on the GPT (Generative Pretrained Transformer) architecture. 
ChatGPT pushed the state-of-the-art for multiple NLP tasks such as language translation, text summarization, sentiment analysis, and question-answering \cite{radford-2019OpenAI:unsupervised_multitask_learners}. 
A major advantage of using ChatGPT is that it requires no training data.
ChatGPT can be instructed via prompts to perform the norm extraction. 

We craft prompts to extract the norms. 
We incrementally refine the prompt by manually inspecting the extracted norms against the expected norms for a few sample clauses from CUAD. 
We found that whereas ChatGPT isn't too sensitive to small changes in a prompt (such as rephrasing or substituting with synonyms), it performs better when given more detailed prompts, i.e., when they contain more information about the concepts to be extracted. 
For instance, we initially did not specify the direction of the relationship between the subject and object for different norm types in the prompt. 
The norms extracted via such prompts often involve the correct parties, i.e., objects and subjects, but in incorrect roles. 
This may be because the roles (directionality) differ depending on the norm type. 
For example, in the case of commitments, the subject commits to the object, whereas in the case of prohibitions, the object prohibits the subject. 
Including the directions explicitly in the definitions for the subject and object in the prompt improved the results by reducing such mismatches. 

As input, we give ChatGPT the prompt followed by the clause from the contract (from which the norm needs to be extracted). 
Below is the prompt we use to extract the norms given a clause.


\begin{chatbox}{ChatGPT Prompt to Extract Norms from Contracts}
\small
\expert{
Given a sentence from a contract (between two parties), extract norms for better contract understanding. 
A norm specifies the expectations from parties involved in the contract in terms of their behavior and actions and is represented by four elements subject, object, antecedent, and consequent. A norm is directed from a subject (the party on whom the norm applies) to an object (the party with respect to whom the norm applies) and is constructed as a conditional relationship involving an antecedent (which brings the norm into force, i.e., the condition on which the action of the subject depends) and a consequent (which brings the norm to satisfaction, i.e., the outcomes of the action). 
A norm can be one of the following four types: }
\newline{}
\prompt{(a) A commitment means that its subject commits to its object to ensure the consequent if the antecedent holds. }
\newline{}
\prompt{(b) A prohibition means that its subject is forbidden by its object from bringing about the consequent if the antecedent holds.}
\newline{}
\prompt{(c) An authorization means that its object authorizes its subject to bring about the consequent if the antecedent holds.}
\newline{}
\prompt{(d) A power means that its subject is empowered by its object to bring about the consequent if the antecedent holds.}
\newline{}
\prompt{The norm types of authorization and power are similar; however, some differences exist. Authorization allows the subject to bring about the consequent (provided it doesn't violate some other norm), whereas power empowers the subject to bring about the consequent when that causes a change in the existing norms.}
\newline{}
\prompt{
In cases where more than one norm exists, extract all norms. Identify the following given a contract sentence.}
\newline{}
\prompt{
1. Norm type: norm type can be `authorization,' `power,' `commitment,' or `prohibition'}
\newline{}
\prompt{
2. Subject: the party on whom the norm applies}
\newline{}
\prompt{
3. Object: the party with respect to whom the norm applies}
\newline{}
\prompt{
4. Antecedent: which brings the norm into force}
\newline{}
\prompt{
5. Consequent: which brings the norm to satisfaction}
\newline{}
\prompt{
The contract sentence is as follows:}
\end{chatbox}

We run our experiments on the ``GPT-3.5-Turbo'' version of ChatGPT. 
Since the ChatGPT model is nondeterministic by default, we set the temperature equal to 0.0 to make it semideterministic and improve reproducibility across runs.

\section{Findings}
\label{sec:Findings}

Example 1 (below) shows the results for successfully extracting correct norms from a sample contract clause using ChatGPT with the above prompt. 
ChatGPT extracts an \emph{authorization} norm and correctly infers all norm elements.

\begin{chatbox}{Example 1: Norm Extraction via ChatGPT}
\small
\expert{[Prompt]}
\newline{}
\prompt{Notwithstanding any other provision of this Agreement, Rogers may terminate this Agreement, at any time, upon sixty (60) days prior written notice to Licensor. 
}
\newline{}
\newline{}
\gpt{}
\prompt{
Norm 1: Authorization}
\newline{}
\prompt{
Subject: Rogers}
\newline{}
\prompt{
Object: Licensor}
\newline{}
\prompt{
Antecedent: Rogers providing sixty (60) days' prior written notice}
\newline{}
\prompt{
Consequent: termination of the Agreement}
\label{box:norm_extraction}
\end{chatbox}

ChatGPT did a good job in extracting norms from contract clauses in many cases.
However, the extractions are not without limitations. 
In some cases, ChatGPT failed to extract a norm that existed in the contract clause, and in other cases, it extracted a norm that did not exist in the clause.
Further, we find that the extracted norms were incorrect sometimes due to incorrect norm type, incorrect norm elements and missing crucial details.

We notice contract clauses containing redactions lead to more incorrect extracted norms.
This implies that contract clauses with redacted information are more challenging for ChatGPT than clauses without redactions. 
This could be critical in situations where some of the information in the contract is redacted due to confidentiality reasons, such as contracts involving national security, and intellectual property.
We leave further anlaysis of norm extraction from clauses containing redactions to future work and only discuss norm extraction from clauses that do not contain any redactions. 

Below we discuss some of the recurring issues in ChatGPT’s norm extractions with examples.


\subsubsection{Incorrect Norm Types}

In some cases, ChatGPT extracts incorrect norm types.
ChatGPT often extracts a commitment norm (in addition to the authorization norm) from clauses that contain only an authorization clause.
Example 2 shows a norm extraction by ChatGPT that inferred an incorrect norm type.
ChatGPT extracts two norms, both have the same norm elements except for the antecedent. 
Whereas norm 1 (authorization) is correct, norm 2 (commitment) is incorrect. 
This is because the original clause uses `may' to describe the potential action making the action optional (has the authority to) but not mandatory (i.e., commitment). 
Hence, the extracted commitment norm is incorrect.
We notice that ChatGPT often confuses what \emph{must} be done with what \emph{can} be done, leading to incorrect norm extraction.

\begin{chatbox}{Example 2: Incorrect Norm Type}
\small
\expert{[Prompt]}
\newline{}
\prompt{
A Party may terminate the JSMA immediately upon the delivery of written notice to the other Party if there has been a Change in Control.}
\newline{}
\newline{}
\gpt{}
\prompt{Norm 1: Authorization}
\newline{}
\prompt{Subject: A Party}
\newline{}
\prompt{Object: the other Party}
\newline{}
\prompt{Antecedent: Change in Control}
\newline{}
\prompt{Consequent: Termination of the JSMA}
\newline{}
\newline{}
\prompt{Norm 2: Commitment}
\newline{}
\prompt{Subject: A Party}
\newline{}
\prompt{Object: the other Party}
\newline{}
\prompt{Antecedent: Change in Control}
\newline{}
\prompt{Consequent: Termination of the JSMA}
\label{box:incorrect_norm_type}
\end{chatbox}

\subsubsection{Incorrect Norm Elements}

In some of the extracted norms, the norm elements are incorrect.
Example 3 shows one such example where multiple extracted norm elements are incorrect.
Note that similar to the previous example (Example 2), ChatGPT extracts an incorrect norm type, i.e., norm 2 (commitment) is incorrect as the original clause contains only authorization. 
In norm 1, the consequent and object are incorrect, and the antecedent is only partially captured.

\begin{chatbox}{Example 3: Incorrect Norm Elements}
\small
\expert{[Prompt]}
\newline{}
\prompt{
Notwithstanding the foregoing, each Party may assign the rights and obligations under this Agreement in whole, without consent of the other Party, to a Third Party or Affiliate in connection with the transfer or sale of all or substantially all of its business or in the event of a merger, consolidation or change in control provided that the assignee assumes in writing and becomes directly obligated to the other Party to perform all of the obligations of assignor under this Agreement.}
\newline{}
\newline{}
\gpt{}
\prompt{Norm 1: Authorization}
\newline{}
\prompt{Subject: Each party}
\newline{}
\prompt{Object: Third Party or Affiliate}
\newline{}
\prompt{Antecedent: Assignment of rights and obligations under the Agreement in connection with the transfer or sale of business, merger, consolidation, or change in control}
\newline{}
\prompt{Consequent: The assignee assumes in writing and becomes directly obligated to perform all obligations of the assignor under the Agreement}
\newline{}
\newline{}
\prompt{Norm 2: Commitment}
\newline{}
\prompt{Subject: Assignee}
\newline{}
\prompt{Object: Other Party}
\newline{}
\prompt{Antecedent: Assignment of rights and obligations under the Agreement in connection with the transfer or sale of business, merger, consolidation, or change in control}
\newline{}
\prompt{Consequent: The assignee becomes directly obligated to perform all obligations of the assignor under the Agreement}
\label{box:incorrect_norm_element}
\end{chatbox}

\subsubsection{Overlooking Crucial Details}

ChatGPT sometimes overlooks crucial details while extracting norms.
For instance, in Example 4, the detail that the agreement will be in effect until March 18\textsuperscript{th} is not captured by any norm (the date is overlooked). 
Further, the detail that 30 days written notice is needed for termination is also lacking from all extracted norms.  

\begin{chatbox}{Example 4: Overlooking Crucial Details}
\small
\expert{[Prompt]}
\newline{}
\prompt{
This Agreement shall be in effect until March 18. 2021, unless sooner terminated by either party upon (30) days written notice, without cause.}
\newline{}
\newline{}
\gpt{}
\prompt{Norm 1: Commitment}
\newline{}
\prompt{Subject: Either party}
\newline{}
\prompt{Object: Either party}
\newline{}
\prompt{Antecedent: Written notice of termination}
\newline{}
\prompt{Consequent: Termination of the Agreement}
\newline{}
\newline{}
\prompt{Norm 2: Authorization}
\newline{}
\prompt{Subject: Either party}
\newline{}
\prompt{Object: Either party}
\newline{}
\prompt{Antecedent: Written notice of termination}
\newline{}
\prompt{Consequent: Termination of the Agreement}
\newline{}
\newline{}
\prompt{Norm 3: Power}
\newline{}
\prompt{Subject: Either party}
\newline{}
\prompt{Object: Either party}
\newline{}
\prompt{Antecedent: Termination of the Agreement}
\newline{}
\prompt{Consequent: End of the Agreement period}
\label{box:overlook_details}
\end{chatbox}

\subsubsection{Hallucinaton}

\emph{Hallucination} refers to a phenomenon where a model generates outputs that are not grounded in the input data.
Hallucination is a known issue in generative models such as ChatGPT and has been found in downstream tasks such as abstractive text summarization.
Hallucination can be highly problematic for norm extraction from contracts that rely heavily on the accuracy of the extracted information, and even minor hallucinations could change the meaning and applicability of clauses.

Example 5 shows a norm extraction by ChatGPT that demonstrates hallucination. 
The subject is inferred as VS (Vendor Service), and although the acronym (VS) is correct (i.e., the entity mentioned in the clauses), its expansion `Vendor Service' is incorrect (the correct expansion is VirtualScopics). 
Interestingly, the input clause doesn't contain the expansion of the acronym, but ChatGPT hallucinated it (the word `Service' occurs in the input clause but not `Vendor'). 
Hallucination could lead to incorrect inferences based on the extracted norms.

Additionally, the norm extraction in Example 5 identifies two norms types for the same extracted norm which is incorrect. 
The commitment norm type is correct and prohibition incorrect.

\begin{chatbox}{Example 5: Hallucination}
\small
\expert{[Prompt]}
\newline{}
\prompt{
VS shall not refer that opportunity to any other third party provider unless PPD declines to bid on such opportunity or does not offer the service in question.}
\newline{}
\newline{}
\gpt{}
\prompt{Norm types: 
- Prohibition, Commitment}
\newline{}
\prompt{Subject: VS (Vendor Services)}
\newline{}
\prompt{Object: Any other third party provider}
\newline{}
\prompt{Antecedent: PPD declines to bid on such opportunity or does not offer the service in question}
\newline{}
\prompt{Consequent: VS shall not refer that opportunity to any other third party provider}

\label{box:hallucination}
\end{chatbox}

\subsubsection{Incorrect Parsing of Conjunctions.}
Clauses with conjunctions, such as AND and OR, are sometimes not parsed correctly, leading to incorrect norm extractions. 
Example 6 shows one such incorrect extraction.
The antecedent (based on the input clause) has two parts joined by a conjunction (`and'), which is incorrectly broken down into antecedent (the first half) and consequent (the second half). 
Hence, the incorrect parsing of conjunction led to the incorrect extraction of norm elements (antecedent and consequent for norm 1 in Example 6).
Further, extracted norm 2 is incorrect as it fails to identify any antecedent, and the extracted consequent is incorrect. 

Whereas in some cases, conjunctions lead to incorrect norm extractions.
In other cases, conjunctions could lead to long and complex antecedents and consequents (when the conjunctions are kept as it is (without rephrasing)).
To address this, we could include instructions in the prompt to break down the conjunctions appropriately. 
However, this complicates the prompt for all cases (even for clauses with no conjunction).
A potential solution could be to have a separate prompt that first breaks down the logical form based on the conjunctions before extracting the norms. 
We leave this to future work as the focus of this study is to identify the challenges and opportunities in norm extraction.

\begin{chatbox}{Example 6: Incorrect Parsing of Conjunctions}
\small
\expert{[Prompt]}
\newline{}
\prompt{
Nothwithstanding the foregoing, this Agreement may be assigned without the AS Parties' consent by MusclePharm in connection with a change of control transaction; provided that the acquirer of MusclePharm shall have financial resources substantially similar or greater than MusclePharm and shall specifically assume the obligations of MusclePharm under this Agreement in writing prior to the consummation of the change of control transaction.}
\newline{}
\newline{}
\gpt{}
\prompt{Norm 1: Authorization}
\newline{}
\prompt{Subject: MusclePharm}
\newline{}
\prompt{Object: AS Parties}
\newline{}
\prompt{Antecedent: The acquirer of MusclePharm shall have financial resources substantially similar or greater than MusclePharm}
\newline{}
\prompt{Consequent: The acquirer of MusclePharm shall specifically assume the obligations of MusclePharm under this Agreement in writing prior to the consummation of the change of control transaction.}
\newline{}
\newline{}
\prompt{Norm 2: Commitment}
\newline{}
\prompt{Subject: MusclePharm}
\newline{}
\prompt{Object: AS Parties}
\newline{}
\prompt{Antecedent: None explicitly stated}
\newline{}
\prompt{Consequent: MusclePharm commits to ensuring that the acquirer of MusclePharm specifically assumes the obligations of MusclePharm under this Agreement in writing prior to the consummation of the change of control transaction.}

\label{box:parsing_conjunction}
\end{chatbox}



\subsubsection{Empty Norm Elements.} For some extracted norms, ChatGPT leaves some norm elements empty (i.e., `', N/A, NULL, or Not Specified).
This is more common for subjects and objects, followed by antecedents.
Consequent is rarely left empty.
Although an empty element in an extracted norm may indicate poor extraction in some cases, it may be the correct response in others. 
For instance, when no precondition (or trigger) exists for a consequent, it implies an empty antecedent. 
Similarly, an empty subject (or object) could mean either \emph{no} party or \emph{all} parties subject to the contract. 
However, in many cases, empty norm elements imply incorrect norm extraction. 
In Example 6, norm 2 of the extracted norms has no antecedent (`None explicitly stated'), implying that the stated consequent would hold without any precondition/trigger, which is incorrect.





\section{Conclusion}
\label{sec:Conclusion}

This study delves into the use of LLMs such as ChatGPT to extract norms from contracts, shedding light on the array of opportunities and challenges inherent in automating contract comprehension. 
We uncover critical insights through a qualitative analysis of the norms extracted by ChatGPT. 

We find that the extraction is influenced by the prompts used. 
A well-crafted prompt that includes definitions of the formal terms used in the prompt yields much better results than less detailed prompts. 
Although a comprehensive prompt consistently yields superior outcomes, the results don't seem too sensitive to minor changes in the prompt (such as rephrasing and substitution with synonyms). 

Despite ChatGPT's overall efficacy in norm extraction, our scrutiny unveils recurrent issues. 
Overlooking crucial information, hallucination, incorrect processing of conjunctions (AND and OR), incorrect norm types, and incorrect norm elements in the extracted norms emerge as persistent challenges. 
Moreover, ChatGPT occasionally identifies norms with empty norm elements, long consequents (that include the subject, object, and antecedents), and incorrect objects and subjects.

Furthermore, our exploration highlights the dearth of high-quality annotated datasets essential for advancing automated contract understanding. 
Existing datasets suffer from limitations such as small size, limited diversity, and subpar annotations, hindering progress in this domain. 

Despite these challenges, our findings underscore the promise of using ChatGPT for norm extraction from unstructured text documents. 
Particularly compelling is the fact that no training or fine-tuning is necessitated, positioning ChatGPT (and similar LLMs) as a preferred model for tasks like contract understanding that lack high-quality annotated datasets. 

In conclusion, this research not only provides valuable insights into the efficacy and challenges of employing LLMs for norm extraction from contracts but also underscores their potential as powerful tools in the realm of automated contract understanding. 


\section*{Acknowledgments}
Thanks to the US NSF (grant IIS-1908374) for support for this research. 


\bibliography{Munindar,Amanul}
\bibliographystyle{splncs04nat}

\end{document}